# PatentNet: A Large-Scale Incomplete Multiview, Multimodal, Multilabel Industrial Goods Image Database

Fangyuan Lei[a,b], Da Huang[a,b], Jianjian Jiang[a,b], Ruijun Ma[c], Senhong Wang[d], Jiangzhong Cao[d], Yusen Lin[a,b], Qingyun Dai[a,b*]

[a] Guangdong Provincial Key Laboratory of Intellectual Property &Big Data, Guangzhou 510665, China

[b] School of Electronic and Information, Guangdong Polytechnic Normal University, Guangzhou 510665, China

[c] Guangdong Industrial Training Center, Guangdong Polytechnic Normal University, Guangzhou 510665, China

[d] School of Information Engineering, Guangdong University of Technology, Guangzhou 510006, China

Corresponding author: Qingyun Dai (1144295091@qq.com)

**Abstract** In deep learning area, large-scale image datasets bring a breakthrough in the success of object recognition and retrieval. Nowadays, as the embodiment of innovation, the diversity of the industrial goods is significantly larger, in which the incomplete multiview, multimodal and multilabel are different from the traditional dataset. In this paper, we introduce an industrial goods dataset, namely PatentNet, with numerous highly diverse, accurate and detailed annotations of industrial goods images, and corresponding texts. In PatentNet, the images and texts are sourced from design patent. Within over 6M images and corresponding texts of industrial goods labeled manually checked by professionals, PatentNet is the first ongoing industrial goods image database whose varieties are wider than industrial goods datasets used previously for benchmarking. PatentNet organizes millions of images into 32 classes and 219 subclasses based on the Locarno Classification Agreement. Through extensive experiments on image classification, image retrieval and incomplete multiview clustering, we demonstrate that our PatentNet is much more diverse, complex, and challenging, enjoying higher potentials than existing industrial image datasets. Furthermore, the characteristics of incomplete multiview, multimodal and multilabel in PatentNet are able to offer unparalleled opportunities in the artificial intelligence community and beyond.

## 1. Introduction

The emergence of deep learning has made breakthrough achievements brought in the context of image object recognition and retrieval[1, 2]. To reach high performance, during training most deep learning models optimize all parameters with the number of labeled images. The availability and quality of the dataset are key factors to affect the performance in many deep learning models.

According to the purpose of image dataset establishment, some datasets are comprehensive, for example the ImageNet[3], COCO [4], and some datasets concentrate on special fields, for example, fashion[5, 6], 3D CAD models[7-9], fault detection[10], face detection[11, 12]. Industrial goods as the embodiment of innovation, there are many varieties. However, the current dataset contains fewer kinds of industrial goods, and does not reflect the diversity of industrial products.

Industrial goods with huge potential value are a symbol of high-quality, convenient, well balanced products. To be successful, the innovative industrial goods are expected to possess desirable characteristics that set them apart from others in a particular branch of industrial. The appearance images are the most direct and vivid way to identify industrial goods. From the perspective of innovation, industrial goods are the fulfillment of creative concepts. And these shapes and structural information constitute the main design elements of the characteristics of its industrial products. For consumers, industrial goods are visually distinguished and identified through shape, texture, structure. However, understanding industrial goods images remains a challenge in real-world applications, because of incomplete view, ambiguity, and discrepancies of industrial goods across domains in a large scale of similar material object and its images.

In general, the appearance characteristics of the industrial goods can be reflected by six views (front view, back view, left side view, right side view, top view, bottom view) and perspective view, as Fig. 1 shown. In practice, view missing often occurs in these seven standard views, as the second row in Fig.1 shown. To the best of our knowledge, most of the existing publicly available benchmark databases for industrial goods object detection, classification and retrieval limit in: 1) fewer categories industrial goods; 2) single-modal industrial goods; 2) single-label image annotation; and 3) complete views data. However, multi-view and multi-modal images associated with the same industrial goods allow for sufficient features of images and thus can better improve image object recognition performance. In addition, industrial goods images usually contain crossing industrial field that must be marked by multiple labels.

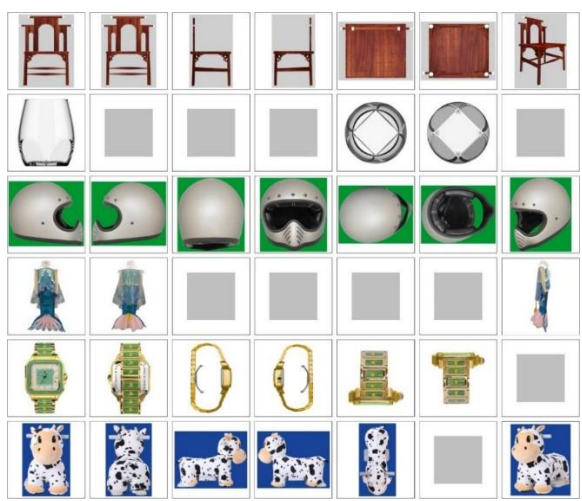

Fig.1 The example of industrial goods. In general, an industrial goods has seven views, i.e., front, back, left side, right side, top, bottom and perspective views. Note that the gray denotes the view missing. The first row is chairs in furnishing; the second row is the kettles in household goods; the third row and the fourth row are the helmets and dresses in articles of clothing and haberdashery respectively; the fifth row is the watches in clocks and watches; and the last row is the cuddly toys in toys.

To address above issues, we propose PatentNet as a large-scale incomplete multiview, multimodal and multilabel benchmark database for industrial goods image recognition, classification, and retrieval. The target of PatentNet is the ImageNet of industrial goods. PatentNet contains 6M images belongs to 32 classes and 219 subclasses, which labeled with multilabel based on Locarno Classification. Fig.2 shows a snapshot of three subclass of garment and furniture. The database will be publicly available at http://iplab.gpnu.edu.cn and GitHub.

The reset of the paper is organized as follows. In section 2, we briefly review the related works on large-scale image datasets. Section 3 describes in detail the collection and setting of the PatentNet database. Section 4 shows the properties of PatentNet. Finally, we conclude the work in Section 5.

## 2. Related works

In this section, we review the large-scale annotated image datasets, which partially or completely contain industrial goods images.

### 2.1 ImageNet

The ImageNet[3] is a large-scale visual database designed for image classification, object localization, visual object detection, scene classification. ImageNet contains more than 20,000 categories in common scenes. With hierarchical structure WordNet, more than 14 million images have been quality-controlled and human-annotated by the project to indicate what objects are pictured and in at least one million of the images, bounding boxes are also provided. ImageNet is an ongoing research effort designed to provide an easily accessible image database for researchers around the world.

### 2.2 COCO dataset

COCO (Common Objects in Context) [4] is a large-scale object detection, segmentation, and captioning dataset. The 2017 version of COCO contains 164K images including 118K for train, 5K for validation, and 20K for test. The images in the dataset are from natural context, which contains common objects in everyday scenes. The dataset contains 80 labeled object categories where the objects are labeled for the precise object localization.

### 2.3 3D dataset

3D ShapeNet[8] is a richly labeled large-scale point cloud dataset for computer graphics, computer vision, robotics. ShapeNet contains 55 common item categories and 513,000 3D models. ModelNet[7] is a 3D CAD dataset for computer vision, computer graphics, robotics and cognitive science. ModelNet contains 127915 3D CAD models from 662 categories, and the ModelNet40 are the uniformly orientated images, which has 12311 models from 40 categories. Although the 3D model can simulate industrial products, they are not a formal industrial goods. PartNet[13] is a large-scale 3D objects dataset for evaluating 3D part recognition. Stanford Cars[14] is a large-scale set of 197 car types for 2D object representations to 3D.

### 2.4 Fashion dataset

Fashion clothing is one of the very import industrial goods, which are specially relation with daily life. Clothing research covers many tasks such as clothing analysis, clothing classification, clothing retrieval, clothing attribute recognition, and clothing understanding. Fashionista[15] is a large-scale annotations dataset for fashion image analysis. The DeepFashion[5] dataset contains 800K annotations fashion image in comprehensive scenes for fashion recognition. For clothing detection, pose estimation and retrieval, DeepFashion2[6] with 873K cloth images is the alternative of DeepFashion. FashionAI[16] is hierarchical dataset with 245 attribute labels, 41 categories, and a total of 357K clothing images. The Fashion-MNIST[17] dataset is a benchmark with 70K 28*28 pixels black and white fashion images.

To summarize, there are some related works on the classification and retrieval of partial industrial goods, as well as the clothing dataset. However, to the best of our knowledge, by far, none of the existing benchmark datasets have established large-scale incomplete multiview, multimodal, multilabel industrial goods images.

## 3. Constructing PatentNet

PatentNet is an ambitious ongoing industrial goods image database. So far, we have constructed 32 class which divide into 219 subclasses containing 6 million industrial goods images. Our goal is to continue to construct around 60 million industrial goods images in the next five years. In this section, we describe the method we use to construct PatentNet.

### 3.1 Data collection

As we all know, the image label is very important for a dataset. To avoid the ambiguity problem caused by random labels, when constructing the industrial image database, we adopt the internationally common classification labels of the design images of industrial products. Therefore, the candidate industrial product images in our database are the design patent images which follow international standard Locarno Classification Agree. The labeling information can be found in the Classifications of the patent document.

The text records the registered information of patent application and legal events, such as publication number, classification number, patent claims. Different multi-view images are the main component of a design patent. The patent right mainly determines the scope of protection through images that reflect the appearance characteristics of the goods.

The candidate industrial data of PatentNet sources from Guangdong Intellectual Property Protection Center of China in 2007 to 2020. We collect candidate data from the Internet by some search engines. The original candidate data is the design patent that has been public published. This candidate data includes bibliographic text and images. The text information contains the category information to which the patent belongs, and the image information reflects the appearance of the industrial goods that the patent applicant wants to protect, as Fig. 2 shown.

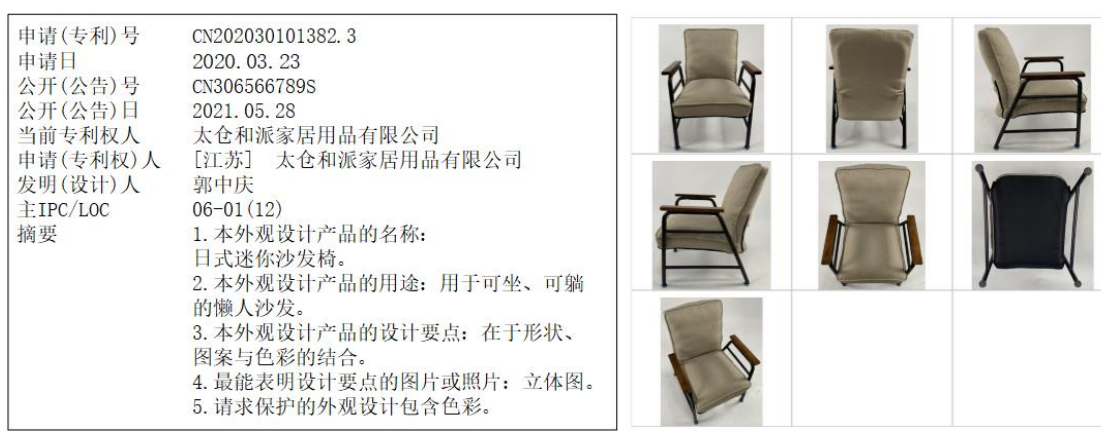

Fig.2. The candidate data include bibliographic text and images.

### 3.2 Data labeling

Data labeling constitutes of hierarchy text and multiview image. In PatentNet, all the information is checked and corrected by patent examiner of the China Intellectual Property Administrator. In this sense, the dataset labeling will be highly accurate.

In the design patent text, field and function information denote the hierarchy class of the industrial goods. Precisely, the field divides into 32 categories, and the function divides into 219 subcategories in total. In some case, the industrial goods may have multiple functions, and this characteristic also obtain from the text. Therefore, the multilabel comprises the multiple function information. The view information of industrial goods images can be extracted from the image through OCR technology.

The naming rules of the image are as follows: AB-CD-EF-GH, where AB represents the field of industrial goods, the value is 0~31. CD is the function of the industrial goods. The range of the CD value is a corresponding relationship between AB, and CD may have multiple value. EF is a unique label, which represents the application number of an industrial goods is assigned by the system of Guangdong Intellectual Property Protection Center of China. GH denotes the view information of an industrial product.

## 4. Properties of PatentNet

PatentNet is constructed upon the hierarchical structure based on Locarno Classification Agreement by World Intellectual Property Organization (WIPO). In the future, PatentNet target is to contain in the order of 50 million cleanly labeled industrial goods images (500-1000 per category). At the time this paper is written, PatentNet consists of 32 classes and 219 categories. Most analysis will be based on the garment class and furniture subclass.

**Scale**   PatentNet aims to provide the most comprehensive and diverse coverage of the industrial goods image in the world. The current 32 fields consist of a total of 5 million cleanly labeled images spread over 219 categories. For each category, we already collect on average over 200 kinds of industrial goods, and over 800 images. Fig.3 shows some distribution of the PatentNet. To the best of our knowledge, PatentNet is already the largest industrial goods database public available for science research, as regards the total number of industrial goods, as well the number of images in each category.

Table 1. Comparison of some of the properties of PatentNet versus other existing datasets

| Dataset | Scale | categories | large-scale | Multi-view | Multi-model | Multi-label | Incomplete | Hierarchy |
|---|---|---|---|---|---|---|---|---|
| ImageNet[3] | 14,197,122 | 21,841 | Y | ✘ | ✘ | Y | ✘ | Y |
| ShapeNet[8] | 3,000,000 | 3,135 | Y | Y | ✘ | ✘ | ✘ | Y |
| ModelNet[7] | 127,915 | 662 | Y | Y | ✘ | ✘ | ✘ | ✘ |
| FashionMNIST[17] | 70,000 | 10 | ✘ | ✘ | ✘ | ✘ | ✘ | ✘ |
| DeepFashion[5] | 289,222 | 50 | Y | ✘ | ✘ | ✘ | ✘ | Y |
| Fashion10000[18] | 1,000,000 | 14 | Y | ✘ | ✘ | ✘ | ✘ | ✘ |
| PartNet[13] | 573,585 | 24 | Y | Y | ✘ | ✘ | ✘ | Y |
| Fashion-Gen[19] | 325,536 | 48 | Y | Y | ✘ | ✘ | ✘ | ✘ |
| PatentNet(our) | 6,000,000 | 219 | Y | Y | Y | Y | Y | Y |

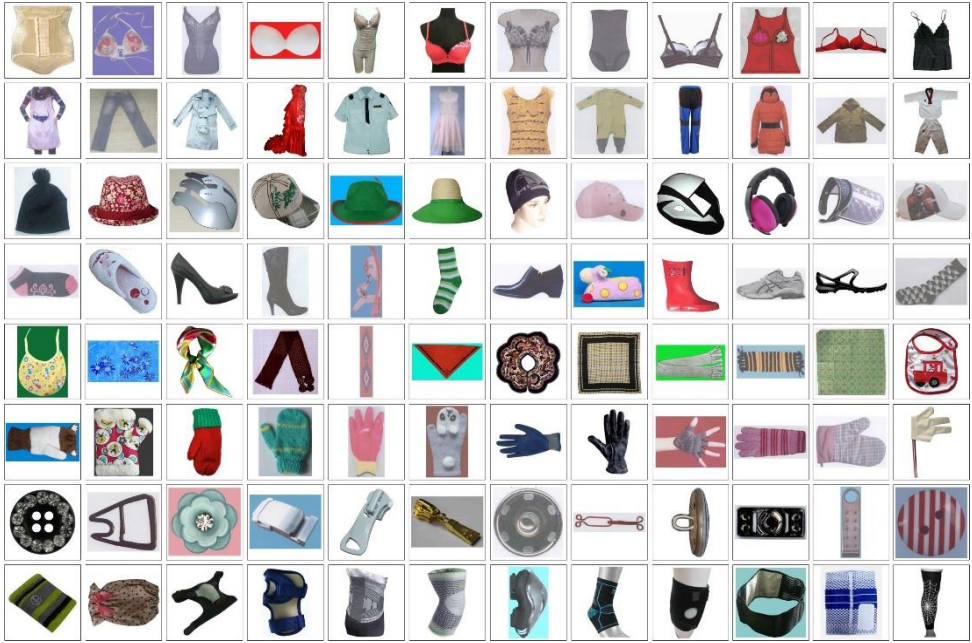

Fig.3. Some examples of PatentNet. The first row is UNDERGARMENTS, LINGERIE, CORSETS, BRASSIÈRES, NIGHTWEAR. The second row is GARMENTS. The third row is HEADWEAR. The fourth row is FOOTWEAR, SOCKS AND STOCKINGS. The fifth row is NECKTIES, SCARVES, NECKERCHIEFS AND HANDKERCHIEFS. The sixth row is GLOVES. The seven row is HABERDASHERY AND CLOTHING ACCESSORIES. The last row is MISCELLANEOUS.

**Hierarchy**   PatentNet organizes the different kinds of industrial goods based on the Locarno Classification standard. The Locarno Classification (LOC) standard consists of three levels, where the top level is the field, and the middle is the function, the last one is the goods name. In most country, only the field and function level are used. Therefore, currently PatentNet has two levels, as Fig. 4 shown. In future, PatentNet will extend to goods name levels.

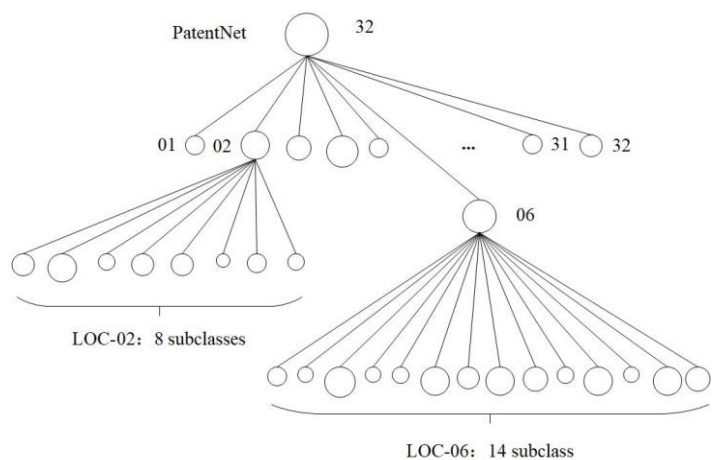

Fig. 4. The hierarchy structure of PatentNet. The PatentNet is divided into two levels. The top level is the field which contains 32 classes, and the second level is the function which total has 219 subclasses. The different field may have different quantity function.

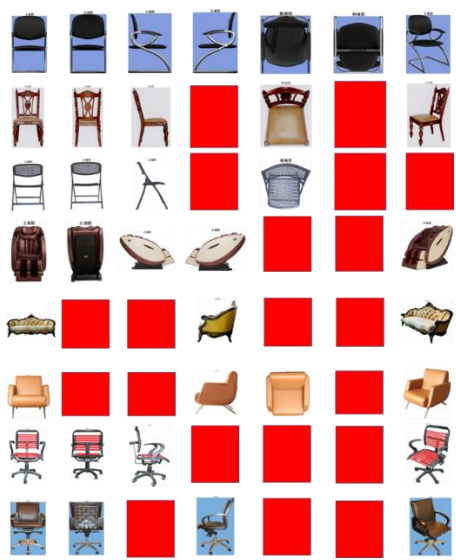

Fig.5. Some examples of incomplete multiview of chair in furniture. The red block represents the missing views. From top to bottom, each row denotes one kind of chair, i.e., bowChair, chairWithoutArmrests, foldingChair, massageChair, multiPersonSofa, singleSofa, swivelChairNonSlipSofa, swivelChairWithSofa, respectively.

**Incomplete Multiview** The image of an industrial goods contains multiple views that reflect the characteristics of the industrial product. These multi-views usually include seven images, which are front view, back view, left side view, right side view, top view, bottom view, and perspective view. The viewing angle gap between these views is more than 90 degrees. As such, there is a weak correlation between the views. In addition, according to the different characteristics of industrial goods, some perspective views are naturally missing. For example, flat objects such as clothing only need a front view and a back view to express the characteristics. Some industrial goods miss certain perspective information that will not affect the characteristics of the industrial products. Consequently, the corresponding view information is also missing. For example, it is similar to the symmetrical goods such as furniture. Fig. 5 shows the incomplete multiview of furniture.

**Multilabel**   PatentNet is organized based on LOC. Some industrial goods have multiple function, therefore the multilabel are provided. In Fig.6, there are some examples of industrial goods with multilabel. One can see that the first row is the lamps with multiple function, which contains bracket lamps, alarm clocks, and recorders. Thus, the industrial goods tend to have three labels: bracket lamps, alarm clocks, and recorders.

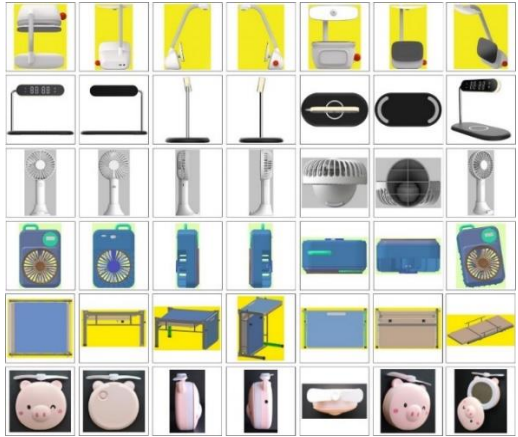

Fig.6. Some examples of multilabel. The labels of the first row are bracket lamps, alarm clocks, and recorders. The second row are the charging apparatus, bracket lamps, and alarm clocks. The third row are fans and supports for telephones. The fourth are fans, alarm clocks, and recorders. The fifth are beds and tables. The last row are fans, mirrors, and lamps.

**Multimodal**   As Fig.2 show, PatentNet not only contains images but also contains texts. The texts are sourced from the registered information of design patent. The text consists of abstract and specification or short specification in the patent. The name and characteristics of the industrial goods are also contained.

## 5. PatentNet Application

In the section, two applications of PatentNet are shown. The first group experiment is the industrial goods image classification and retrieval. The last experiment is the goods image clustering, which focus on the effect of the incomplete multiview data.

### 5.1 industrial goods classification and retrieval

In these experiments, we use the fashion images which belong to a subset of the PatentNet. The fashion subset has 8 subclasses, and total 48000 images, as Fig. 3 shown.

In the classification task, we use 40000 images for training, and 8000 images for testing. The top-1 accuracy of classification is listed in Table 2. To evaluate the performance of our PatentNet in classification task, we conduct classification experiment by using some classical classification methods, including VGG19[20], GoogLeNet[21], Resnet50[22], Densenet121[23] and Efficientnet-B0[24]. As in Table 2, the experimental results show that the Top-1 classification accuracy is lower. Meanwhile, in the ImageNet the Efficientnet-B0 achieves 76.3%[24], which is almost 14% over than in PatentNet. Therefore, the PatentNet is more challenging.

In this retrieval task, we use the pre-training VGG19[20] and Resnet152[22] models for images feature extraction. Euclidean distance is then leveraged as a similarity measure to get the similar industrial goods images in the querying. On the dataset split setting, we utilize 40000 images to build the feature database and the rest 8000 images for the querying testing.

We select some retrieval methods such as CSQ[25] and HashNet[26], which work well on other datasets in retrieval tasks. In the evaluation, we adopt the mean average precision(mAP) denoted the average of AP on all queries as the evaluation metrics, which are widely used in retrieval tasks. Table 3. presents the top-10 mAP of the retrieval, and the category level experimental result shows in Fig.7. The retrieval task indicate that the category imbalance will affect the retrieval accuracy. The retrieval task also demonstrates that the proposed database PatentNet is more challenging, and the corresponding algorithms need to be improved and developed.

Table 2. Top-1 Accuracy of Classification

| Models | ACC |
|---|---|
| Resnet-50[22] | 63.25% |
| VGG-19[20] | 72.95% |
| Densenet-121[23] | 71.57% |
| GoogLeNet[21] | 69.60% |
| Efficientnet-B0[24] | 62.92% |

Table 3. Top-10 mAP of Retrieval Task

| Models | mAP |
|---|---|
| Resnet-152[22] | 72.05% |
| VGG-19[20] | 66.96% |
| CSQ[25] | 54.00% |
| HashNet[26] | 26.10% |

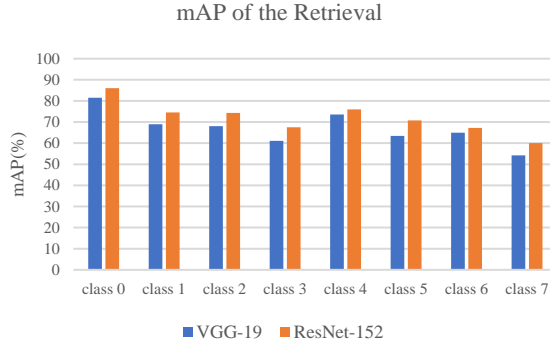

Fig 7. The Top-10 mAP of retrieval on category level

### 5.2 Clustering on incomplete multiview

Compared to other available datasets, PatentNet provides image data with incomplete multiview. In general, one industrial goods can be described by at most 7 views that represent images, which are front view, back view, left side view, right side view, top view, bottom view, and perspective view, but they are often missing. Many possible methods could be applied to employ the missing information.

We collect 400 chairs with 2159 images, which are categorized into eight subclasses: bowChair, chairWithoutArmrests, foldingChair, massageChair, multiPersonSofa, singleSofa, swivelChairNonSlipSofa, swivelChairWithSofa, as Fig.5 shown. Each subclass has 50 chairs, and some of them belongs incomplete multiview. We utilize VGG16[20] to obtain 4096 dimension deep feature for each image. Table 4 shows the statistical characteristic of incomplete multiview dataset.

In this experiment, we demonstrate the usefulness of the PatentNet multi-view against some state-of-the-art incomplete multi-view clustering methods. Thus, we choose Spectral Clustering (SC) [27], IMG[28], DAIMC[29], IMSC_AGL[30] and APMC[31], where Spectral Clustering are performed on each view. To further prove the rational use of multi-view data can improve clustering performance, we concatenate all views into a single view with long dimensions, and then run SC to obtain the clustering results. Since SC cannot directly deal with incomplete multi-view data, we first fill the missing information with average feature values in the corresponding view, and then perform SC. Empirically, the number of nearest neighbors is accounted for 10% of the dataset size for SC. We select the best two views from the industrial goods dataset as the input of IMG, because it cannot work on more than two-view scenario. Since it is very complicated to adjust the order of multi-view data for APMC when processing more than 3 views, we select the best three views. What's more, in order to make the experiments fair enough, we set the parameters of each method just as reported in the corresponding papers. To randomize the experiment, we run each method for 20 times to record the mean performance as well as the standard deviations for every method.

**Table 4.** Statistical characteristic of incomplete multiview datasets

| Dataset | Samples | Clusters | Views | Observed Samples | Features Dimension |
|---|---|---|---|---|---|
| chair subset | 400 | 8 | 7 | 400, 340, 325, 339, 296, 71, 388 | 4096, 4096, …, 4096 |

There are many criteria to evaluate the clustering performances[32]. In our experiments, we choose three most widely used evaluation metrics, namely, clustering accuracy (ACC), normalized mutual information (NMI), and Purity to conduct comprehensively evaluation. These criteria can be calculated by comparing the obtained clustering labels of each sample with the ground-truth labels provided by the dataset. For the three kind of evaluation metrics, a higher value indicates the better performances. The readers can refer to this paper[33] to get more details about their definitions.

**Table 5.** ACC (%), NMI (%), Purity (%) of different incomplete multiview clustering methods on the chair subset of PatentNet. In SC(i), the parameter i denotes the views number.

| Method | ACC (%) | NMI (%) | Purity (%) |
|---|---|---|---|
| SC(1) | 74.01±0.40 | 70.06±0.31 | 74.01±0.40 |
| SC(2) | 28.16±2.12 | 15.37±1.77 | 29.61±2.42 |
| SC(3) | 36.83±2.04 | 24.32±1.40 | 38.04±1.68 |
| SC(4) | 29.29±2.80 | 16.81±2.16 | 30.60±2.71 |
| SC(5) | 20.82±1.71 | 9.55±3.07 | 21.47±1.73 |
| SC(6) | 23.31±0.45 | 7.67±0.28 | 24.70±0.52 |
| SC(7) | 78.71±0.17 | 68.98±0.24 | 78.71±0.17 |
| SC(concate) | 83.49±0.06 | 77.36±0.04 | 83.49±0.06 |
| IMG(V1-V7) | 70.60±3.71 | 70.58±1.26 | 73.49±1.55 |
| DAIMC | 83.58±0.12 | 79.55±0.11 | 83.58±0.12 |
| APMC(V1-V3-V7) | 84.74±0.06 | 79.22±0.12 | 84.74±0.06 |
| IMSC_AGL | 83.60±0.17 | 81.08±0.28 | 83.60±0.17 |

The ACC, NMI and Purity of above different methods on the chair subset of PatentNet are enumerated in Table 5. Analysis of the data, we can get some observations as following: (1) the clustering performances are quite different in single view. This is mainly because each view has difference in the feature scales and distributions. (2) Most of the above incomplete multi-view clustering methods can achieve much better performance than the Single-View. Therefore, it is worthy to study how to appropriately combine multiple views to improve the clustering performance. Moreover, this experiment proves that the PatentNet can be used to evaluate the incomplete multi-view clustering algorithms.

## 6. Conclusions and Future Work

### 6.1 Ongoing PatentNet

The current PatentNet contains various industrial goods sourced from Guangdong Intellectual Property Protection Center of China in 2007 to 2020. As we know, the number of industrial goods increases 700k per year. Therefore, extending the quantity of the PatentNet is our first goal. In future, we also want to enlarge the PatentNet to coverage more country and region, such as Japanese, U.S, Hong Kong. Meanwhile, the PatentNet covers industrial goods only from the Intellectual Property Administration. We further plan to extend the industrial goods images of PatentNet from an e-commerce platform in the future.

### 6.2 Exploiting PatentNet

We have the purpose to make the PatentNet be one of main dataset for a broad of range of artificial intelligence related research. In the further, we will build more potential application scenarios related to PatentNet, and will take one step forward to explore more applications in future.

**A benchmark dataset**. Deep learning has become the universal technique in computer vision, nature language process, etc. Deep learning models possess excellent performance in target identification and image classification. The large-scale datasets play critical role in model training of AI, especially in deep learning, such as ImageNet, MS COCO. As a large-scale industrial goods dataset, PatentNet will be a new alternative and more challenging benchmark dataset for future research.

**Incomplete multiview dataset**. The incompleteness of data and the imbalance between data categories are common problems. In the PatentNet database, the lack of view data of industrial goods and the difference quantity among the industrial goods categories provide a concrete real sample case for this type of research. PatentNet is the first large-scale dataset with weak correlations between perspectives. It shows that the gap between the perspectives of industrial goods is large, and the correlation between the views is relatively weak. With this kind of association, traditional machine learning methods cannot effectively discover the large-span weakly associations. It is necessary to develop a novel algorithm to solve such problems.

**Multimodal knowledge graph**. In the PatentNet dataset, the image represents given industrial goods, and the corresponding text contains the description information of the industrial goods and related legal events. There is a potential correlation between the images of industrial products. Therefore, it worth constructing a multimodal knowledge graph centered on industrial products. The cognitive intelligence technique will introduce to find the relationship within the multimodal knowledge graph of industrial goods.

**Semantic concepts and computer vision research**. PatentNet has hierarchical structure and full coverage of the industrial goods image may help advance the understanding between 3-D projection of computer and the human vision. For example, it is a very complicated process to full query the multiview images of the industrial goods by the name of given industrial goods. It is worthy to explore the corresponding between the conceptual level of the product name and the actual product. At the same time, further automatic fined classification of industrial product images according to names is an area that needs further exploration.

**A bridge between innovation and protection**. In PatentNet, it is deserving of paying attention to assigning new industrial goods images to corresponding categories. Furthermore, partial design patent protection law will implement, and comparing the differences of the partial features of images will have practical significance. Promoting innovation is very important to construct the cross-modal relationship between industrial goods and the actual products selling on the e-commerce platform.

ACKNOWLEDGMENTS

This work was supported in part by the National Natural Science Foundation of China under Grant U1701266, in part by Guangdong Provincial Key Laboratory of Intellectual Property and Big Data under Grant 2018B030322016, in part by Special Projects for Key Fields in Higher Education of Guangdong under Grant 2020ZDZX3077, and in part by Qingyuan Science and Technology Plan Project under Grant 170809111721249 and Grant 170802171710591.

**Reference**

[1] Y. Guo, Y. Liu, A. Oerlemans, S. Lao, S. Wu, M.S. Lew, Deep learning for visual understanding: A review, Neurocomputing, 187 (2016) 27-48.

[2] W. Rawat, Z. Wang, Deep convolutional neural networks for image classification: A comprehensive review, Neural computation, 29 (2017) 2352-2449.

[3] J. Deng, W. Dong, R. Socher, L.-J. Li, K. Li, L. Fei-Fei, Imagenet: A large-scale hierarchical image database, 2009 IEEE conference on computer vision and pattern recognition, Ieee, 2009, pp. 248-255.

[4] T.-Y. Lin, M. Maire, S. Belongie, J. Hays, P. Perona, D. Ramanan, P. Dollár, C.L. Zitnick, Microsoft coco: Common objects in context, European conference on computer vision, Springer, 2014, pp. 740-755.

[5] Z. Liu, P. Luo, S. Qiu, X. Wang, X. Tang, Deepfashion: Powering robust clothes recognition and retrieval with rich annotations, Proceedings of the IEEE conference on computer vision and pattern recognition, 2016, pp. 1096-1104.

[6] Y. Ge, R. Zhang, X. Wang, X. Tang, P. Luo, Deepfashion2: A versatile benchmark for detection, pose estimation, segmentation and re-identification of clothing images, Proceedings of the IEEE/CVF Conference on Computer Vision and Pattern Recognition, 2019, pp. 5337-5345.

[7] Z. Wu, S. Song, A. Khosla, F. Yu, L. Zhang, X. Tang, J. Xiao, 3d shapenets: A deep representation for volumetric shapes, Proceedings of the IEEE conference on computer vision and pattern recognition, 2015, pp. 1912-1920.

[8] A.X. Chang, T. Funkhouser, L. Guibas, P. Hanrahan, Q. Huang, Z. Li, S. Savarese, M. Savva, S. Song, H. Su, Shapenet: An information-rich 3d model repository, arXiv preprint arXiv:1512.03012, (2015).

[9] M. Aubry, D. Maturana, A.A. Efros, B.C. Russell, J. Sivic, Seeing 3d chairs: exemplar part-based 2d-3d alignment using a large dataset of cad models, Proceedings of the IEEE conference on computer vision and pattern recognition, 2014, pp. 3762-3769.

[10] B. de Bruijn, T.A. Nguyen, D. Bucur, K. Tei, Benchmark Datasets for Fault Detection and Classification in Sensor Data, SENSORNETS, 2016, pp. 185-195.

[11] T. Karras, S. Laine, T. Aila, A style-based generator architecture for generative adversarial networks, Proceedings of the IEEE/CVF Conference on Computer Vision and Pattern Recognition, 2019, pp. 4401-4410.

[12] R. Vemulapalli, A. Agarwala, A compact embedding for facial expression similarity, Proceedings of the IEEE/CVF Conference on Computer Vision and Pattern Recognition, 2019, pp. 5683-5692.

[13] K. Mo, S. Zhu, A.X. Chang, L. Yi, S. Tripathi, L.J. Guibas, H. Su, Partnet: A large-scale benchmark for fine-grained and hierarchical part-level 3d object understanding, Proceedings of the IEEE/CVF Conference on Computer Vision and Pattern Recognition, 2019, pp. 909-918.

[14] J. Krause, M. Stark, J. Deng, L. Fei-Fei, 3d object representations for fine-grained categorization, Proceedings of the IEEE international conference on computer vision workshops, 2013, pp. 554-561.

[15] R. He, C. Lin, J. McAuley, Fashionista: A fashion-aware graphical system for exploring visually similar items, Proceedings of the 25th International Conference Companion on World Wide Web, 2016, pp. 199-202.

[16] X. Zou, X. Kong, W. Wong, C. Wang, Y. Liu, Y. Cao, Fashionai: A hierarchical dataset for fashion understanding, Proceedings of the IEEE/CVF Conference on Computer Vision and Pattern Recognition Workshops, 2019, pp. 0-0.

[17] H. Xiao, K. Rasul, R. Vollgraf, Fashion-mnist: a novel image dataset for benchmarking machine learning algorithms, arXiv preprint arXiv:1708.07747, (2017).

[18] B. Loni, L.Y. Cheung, M. Riegler, A. Bozzon, L. Gottlieb, M. Larson, Fashion 10000: an enriched social image dataset for fashion and clothing, Proceedings of the 5th acm multimedia systems conference, 2014, pp. 41-46.

[19] N. Rostamzadeh, S. Hosseini, T. Boquet, W. Stokowiec, Y. Zhang, C. Jauvin, C. Pal, Fashion-gen: The generative fashion dataset and challenge, arXiv preprint arXiv:1806.08317, (2018).

[20] K. Simonyan, A. Zisserman, Very deep convolutional networks for large-scale image recognition, arXiv preprint arXiv:1409.1556, (2014).

[21] C. Szegedy, W. Liu, Y. Jia, P. Sermanet, S. Reed, D. Anguelov, D. Erhan, V. Vanhoucke, A. Rabinovich, Going deeper with convolutions, Proceedings of the IEEE conference on computer vision and pattern recognition, 2015, pp. 1-9.

[22] K. He, X. Zhang, S. Ren, J. Sun, Deep residual learning for image recognition, Proceedings of the IEEE conference on computer vision and pattern recognition, 2016, pp. 770-778.

[23] G. Huang, Z. Liu, L. Van Der Maaten, K.Q. Weinberger, Densely connected convolutional networks, Proceedings of the IEEE conference on computer vision and pattern recognition, 2017, pp. 4700-4708.

[24] M. Tan, Q. Le, Efficientnet: Rethinking model scaling for convolutional neural networks, International Conference on Machine Learning, PMLR, 2019, pp. 6105-6114.

[25] L. Yuan, T. Wang, X. Zhang, F.E. Tay, Z. Jie, W. Liu, J. Feng, Central similarity quantization for efficient image and video retrieval, Proceedings of the IEEE/CVF Conference on Computer Vision and Pattern Recognition, 2020, pp. 3083-3092.

[26] Z. Cao, M. Long, J. Wang, P.S. Yu, Hashnet: Deep learning to hash by continuation, Proceedings of the IEEE international conference on computer vision, 2017, pp. 5608-5617.

[27] A. Ng, M. Jordan, Y. Weiss, On Spectral Clustering: Analysis and an algorithm, in: T. Dietterich, S. Becker, Z. Ghahramani (Eds.) Advances in Neural Information Processing Systems, MIT Press, 2002.

[28] H. Zhao, H. Liu, Y. Fu, Incomplete multi-modal visual data grouping, Proceedings of the Twenty-Fifth International Joint Conference on Artificial Intelligence, 2016, pp. 2392-2398.

[29] M. Hu, S. Chen, Doubly Aligned Incomplete Multi-view Clustering, International Joint Conference on Artificial Intelligence, 2018, pp. 2262-2268.

[30] J. Wen, Y. Xu, H. Liu, Incomplete Multiview Spectral Clustering With Adaptive Graph Learning, IEEE Transactions on Cybernetics, 50 (2020) 1418-1429.

[31] J. Guo, J. Ye, Anchors bring ease: An embarrassingly simple approach to partial multi-view clustering, Proceedings of the AAAI Conference on Artificial Intelligence, 2019, pp. 118-125.

[32] E. Amigó, J. Gonzalo, J. Artiles, F. Verdejo, A comparison of extrinsic clustering evaluation metrics based on formal constraints, Information retrieval, 12 (2009) 461-486.

[33] K. Zhan, X. Chang, J. Guan, L. Chen, Z. Ma, Y. Yang, Adaptive structure discovery for multimedia analysis using multiple features, IEEE transactions on cybernetics, 49 (2018) 1826-1834.